\documentclass[10pt,twocolumn,letterpaper]{article}

\usepackage[T1]{fontenc}
\usepackage[utf8]{inputenc}
\usepackage[margin=0.75in,columnsep=0.25in]{geometry}
\usepackage{amsmath,amssymb}
\usepackage{newtxtext,newtxmath}
\usepackage{microtype}
\usepackage{graphicx}
\usepackage[table]{xcolor}
\usepackage{booktabs,multirow,tabularx}
\usepackage{adjustbox}
\usepackage{algorithm,algorithmic}
\usepackage{needspace}
\usepackage{titlesec}
\usepackage[font=small,labelfont=bf,labelsep=period]{caption}
\usepackage[numbers,sort&compress]{natbib}
\usepackage{xurl}
\usepackage[hidelinks,pdfencoding=auto]{hyperref}

\titleformat{\section}{\large\bfseries}{\thesection}{0.6em}{}
\titleformat{\subsection}{\normalsize\bfseries}{\thesubsection}{0.6em}{}
\titleformat{\subsubsection}[runin]{\normalsize\bfseries}{\thesubsubsection}{0.6em}{}[.]
\titleformat{\paragraph}[runin]{\normalsize\bfseries}{}{0pt}{}[.]
\titlespacing*{\section}{0pt}{10pt plus 2pt minus 2pt}{4pt}
\titlespacing*{\subsection}{0pt}{8pt plus 2pt minus 2pt}{3pt}
\titlespacing*{\subsubsection}{0pt}{6pt plus 1pt minus 1pt}{0.5em}
\titlespacing*{\paragraph}{0pt}{6pt plus 1pt minus 1pt}{0.5em}
\renewenvironment{abstract}
  {\section*{Abstract}}
  {\par\vspace{5pt}}

\pdftrailerid{}
\input{sections/algorithm-environment}

\title{Rethinking Generalization in Knowledge Graph Embedding:\\
An Empirical Study and Sparse Peak Regularization}
\author{\textbf{Linyu Li}\textsuperscript{1,2},
  \textbf{Zhi Jin}\textsuperscript{1,2*},
  \textbf{Yuanpeng He}\textsuperscript{1,2},
  \textbf{Dongming Jin}\textsuperscript{1,2},
  \textbf{Haoran Duan}\textsuperscript{1,2},
  \textbf{Zhengwei Tao}\textsuperscript{1,2},\\[1pt]
  \textbf{Xuan Zhang}\textsuperscript{1,2},
  \textbf{Jiandong Li}\textsuperscript{1,2},
  \textbf{Huanyao Zhang}\textsuperscript{1,2},
  \textbf{Gadeng Luosang}\textsuperscript{3},
  \textbf{Nyima Tashi}\textsuperscript{3}\\[5pt]
  {\fontsize{10}{12}\selectfont
  \textsuperscript{1}School of Computer Science, Peking University, Beijing 100871, China\\
  \textsuperscript{2}Key Laboratory of High Confidence Software Technologies (PKU), Ministry of Education\\
  \textsuperscript{3}School of Information Science and Technology, Tibet University, Lhasa 850000, China\\[2pt]
  \textsuperscript{*}Corresponding author\\
  \href{mailto:linyuli@stu.pku.edu.cn}{linyuli@stu.pku.edu.cn},
  \href{mailto:zhijin@pku.edu.cn}{zhijin@pku.edu.cn}}}
\date{}

\makeatletter
\renewcommand{\@maketitle}{\begingroup
  \centering
  {\fontsize{17}{20}\selectfont\bfseries\@title\par}
  \vskip 9pt
  {\fontsize{11}{13}\selectfont\@author\par}
  \endgroup
  \vskip 10pt
}
\makeatother

\hypersetup{
  pdftitle={Rethinking Generalization in Knowledge Graph Embedding: An Empirical Study and Sparse Peak Regularization},
  pdfauthor={Linyu Li, Zhi Jin, Yuanpeng He, Dongming Jin, Haoran Duan, Zhengwei Tao, Xuan Zhang, Jiandong Li, Huanyao Zhang, Gadeng Luosang, Nyima Tashi},
  pdfsubject={Knowledge graph embedding and regularization},
  pdfkeywords={Knowledge Graph, Knowledge Graph Embedding, Link Prediction, Regularization},
  pdfcreator={},
  pdfproducer={}
}

\begin{document}
\maketitle
\begin{abstract}
Knowledge graphs (KGs) provide structured representations of entities and relations, and underpin data management and Web applications such as search, recommendation, and question answering. Since Web-scale KGs such as Freebase and Wikidata are inherently incomplete, knowledge graph embedding (KGE) maps entities and relations into continuous vector spaces and is widely used to infer missing facts through link prediction. Despite extensive progress in KGE model design, the role of regularization has been less systematically examined, especially beyond tensor-decomposition models. The primary contribution of this paper is a comprehensive empirical study of regularization for KGE from the perspective of generalization. Across translation-based, tensor-decomposition-based, GNN-based, temporal, and multimodal KGE models, we show that overfitting is common across model families and that appropriate regularization can substantially improve both generalization and predictive performance. Motivated by these findings, we further propose Sparse Peak Regularization (SPR), a simple and model-agnostic regularizer that introduces rank-based selective sparsity into the regularization target. SPR concentrates the penalty on dominant high-energy embedding components, which contribute most to the embedding norm and overfitting risk, while leaving low-energy components unpenalized so that weak but useful signals are not unnecessarily suppressed. Experiments on eight benchmark datasets and diverse KGE models show that SPR achieves the best results in nearly all settings compared with existing regularizers such as N3, DURA, and ER. These results show that regularization deserves more systematic attention in KGE, and that selective regularization is an effective and broadly applicable strategy for improving the generalization of KGE models.
\end{abstract}

\noindent\textbf{Keywords:} Knowledge Graph; Knowledge Graph Embedding; Link Prediction; Regularization.

\section{Introduction}
\begin{figure}[t]
    \centering
    \includegraphics[width=\linewidth]{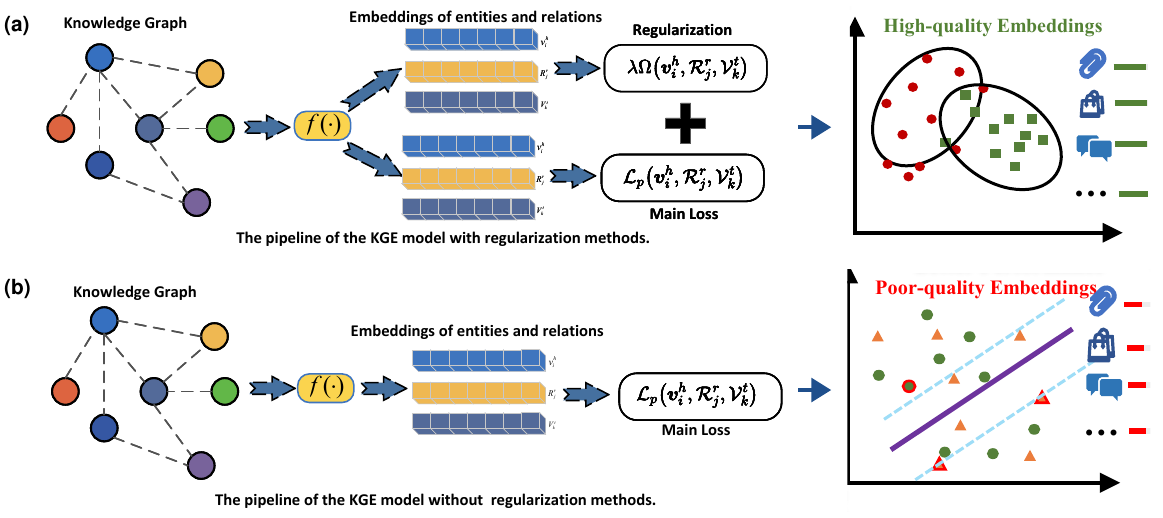}
    \caption{Overview of regularization and generalization in KGE. (a) Regularized training combines the main prediction loss $\mathcal{L}_p$ with a weighted penalty $\lambda\Omega$, encouraging better-separated embeddings and stronger downstream behavior. (b) Without regularization, KGE minimizes only $\mathcal{L}_p$ and is more susceptible to poorly separated embeddings and weaker downstream performance.}
    \label{fig:overview}
\end{figure}

Knowledge graphs (KGs) \cite{liang2024survey,ji2021survey} are graph-structured representations that model the real world.
As structured domain knowledge, KGs support ontology-based data access \cite{calvanese2017ontop}, large language models (LLMs) \cite{pan2024unifying}, and recommender systems \cite{guo2020survey}. Document-based retrieval-augmented generation also integrates external knowledge into LLMs \cite{dong2025decoupling}.
Representative Web-scale resources such as Freebase and Wikidata, both built collaboratively on the Web, organize knowledge symbolically through entities and relations \cite{bollacker2008freebase,vrandevcic2014wikidata}. Such discrete structures are difficult to feed directly into modern learning systems. Therefore, learning effective Knowledge Graph Embedding (KGE) models, which represent entities and relations as low-dimensional vectors and underpin downstream tasks such as link prediction, is crucial for enhancing the usability of KGs. In practice, the quality of the learned embeddings has a large impact on these downstream tasks. Figure~\ref{fig:overview} connects this consequence to the training objective: adding a weighted regularization term to the main prediction loss (panel (a)) encourages better-separated representations and stronger downstream behavior, whereas a model trained only with the main loss (panel (b)) is more susceptible to poorly separated embeddings and weak generalization. Learning KGE models that generalize well is therefore a key step towards the practical use of KGs. This is particularly relevant to Web knowledge completion: Web-scale KGs are inherently incomplete, and KGE-based link prediction infers missing facts from the observed entity--relation graph, so embeddings that generalize beyond the training triples benefit the Web applications built on these KGs.

\begin{figure*}[t]
    \centering
    \includegraphics[width=0.94\linewidth]{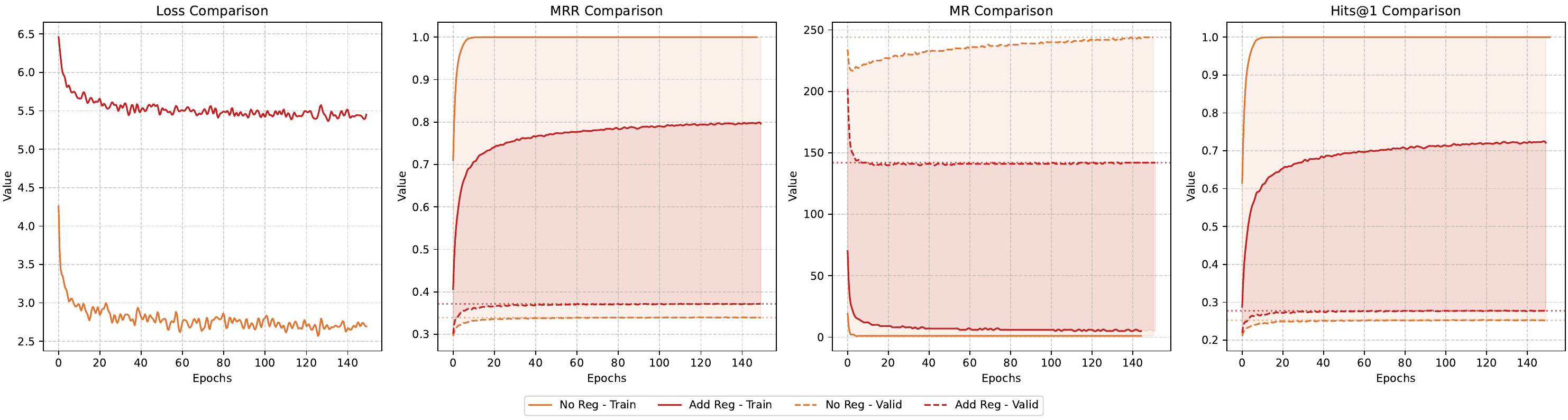}
    \caption{Comparison of the overfitting performance of the GIE \cite{cao2022geometry} model without regularization and with regularization training on the FB15K-237 dataset: Visual comparison of the model Loss, MRR, MR, and Hits@1 metrics as they change with Epoch. The shaded region denotes the train-validation gap. By adding the regularization method, this generalization gap is substantially narrowed.}
    \label{fig:1}
\end{figure*}
Recent studies have developed a wide range of KGE models \cite{bordes2013translating,liang2024clustering,liang2023knowledge,li2025multi,cao2026cakge,wang2024large,yao2019kg,li2026complete,xiao2023knowledge,badreddine2025breaking}. However, most current studies overlook the overfitting problem of KGE models in practical applications: the model learns the noise and specific relation patterns in the training set and performs poorly on new, unseen data, which limits its generalization ability and achievable performance. In addition, unlike large language models, which are trained on massive corpora with only a few passes over the data, a KGE model is trained on a comparatively small knowledge graph and is usually optimized for at least 100 epochs. Repeatedly fitting such limited data over so many epochs makes overfitting particularly common, which is an important reason why regularization deserves more attention in KGE.

To better understand and ultimately solve the overfitting problem in KGE, this paper conducts an extensive empirical study on regularization for KGE models. Our analysis covers various types of KGE models, especially translation-based, tensor decomposition-based, and neural network-based models. Through extensive experiments on these models and multiple KG datasets, we find that many existing KGE models overfit, as evidenced by a large discrepancy between their high training performance and low validation performance. Taking Figure~\ref{fig:1} as an example, the GIE \cite{cao2022geometry} model quickly achieves nearly perfect training accuracy without regularization but performs much worse on the validation set, a clear sign of overfitting. Moreover, without regularization, its MR gradually increases, which means that correct entities are ranked lower on average and further illustrates poor generalization. In contrast, when regularization is added, convergence slows down and more training epochs are required. Although the training performance decreases slightly, the validation performance improves significantly, which is exactly what better generalization requires. Similarly, Table~\ref{tab:kgc-main-dataset} shows that adding regularization improves test performance over the corresponding unregularized baseline for every evaluated backbone. Further empirical results are reported in Section~\ref{sec:experiment}.

Existing regularization methods for KGE often constrain model complexity through embedding-vector norms. For example, the Frobenius norm regularizer \cite{nickel2011three,yang2014embedding} is widely used in various models due to its simplicity. Subsequent regularizers such as N3 \cite{lacroix2018canonical}, DURA \cite{zhang2020duality,wang2022duality}, ER \cite{cao2022er}, and VIR \cite{xiao2024knowledge} achieve good results, but they were designed for and evaluated mainly on KGE models based on tensor decomposition. We propose a KGE regularization method called Sparse Peak Regularization (SPR). Its lightweight design supports multiple KGE model families, including GNN-based, translation-based, and tensor-\allowbreak decomposition-based models, and it achieves the best results in nearly all of our experiments. In summary, our contributions are threefold:

\begingroup
\setlength{\parskip}{0pt}
\par (1) The primary contribution of this paper is an extensive empirical study of regularization for KGE. Whereas prior regularization research was validated almost only on tensor-decomposition models, we show that overfitting is in fact a widespread but long-overlooked problem across translation-based, tensor-\allowbreak decomposition-based, GNN-based, temporal, and multimodal KGE models, and that a proper use of regularization consistently alleviates it, improving generalization and performance relative to the corresponding unregularized baselines.\par
\par (2) We propose a simple and efficient regularization method, SPR, for the KGE model, which uses the idea of sparsity to focus the regularization penalty on the dominant components while leaving the many unimportant ones out of it.\par
\par (3) On eight datasets and various KGE models, SPR achieves the best results in nearly all settings, and our theoretical analysis characterizes how it differs from Dropout and uniform L2 penalties.\par
\endgroup

\section{Related Work}

\paragraph{KGE models} Translation-based (TB) models \cite{cao2024knowledge} treat each relation as a translation or geometric transformation between entity vectors, as in TransE \cite{bordes2013translating}, TransH \cite{wang2014knowledge}, TransR \cite{lin2015learning}, RotatE \cite{sun2019rotate}, HAKE \cite{zhang2020learning}, GIE \cite{cao2022geometry}, ROTH \cite{chami2020low}, and ConE \cite{bai2021modeling}. Tensor-decomposition-based (TDB) models factorize the binary KG tensor; DistMult \cite{yang2014embedding}, ComplEx \cite{trouillon2016complex}, SimplE \cite{kazemi2018simple}, RESCAL \cite{nickel2011three}, and TuckER \cite{balavzevic2019tucker} are closely related to the CP decomposition \cite{hitchcock1927expression}.

\paragraph{Regularization for KGE} Regularization \cite{hastie2009elements} penalizes model complexity to keep the model from fitting noise in the training data. Besides the Frobenius norm \cite{nickel2011three,yang2014embedding}, KGE regularizers such as N3 \cite{lacroix2018canonical}, DURA \cite{zhang2020duality,wang2022duality}, ER \cite{cao2022er}, and VIR \cite{xiao2024knowledge} have mainly been designed for and evaluated on TDB models. Data augmentation \cite{xie2023enhance,shomer2023degree,chang2023counterfactual}, contrastive learning \cite{tan2023kracl,zhang2024hasa}, and non-sampling training \cite{li2021nonsampling} improve other parts of learning and are complementary to SPR. Since we find serious overfitting well beyond TDB models, we design SPR to apply across model families.

\paragraph{Empirical analyses of KGE} Prior studies analyze GCN-based completion \cite{zhang2022rethinkinggcn}, evaluation protocols \cite{tiwari2021revisiting}, embedding reliability \cite{egger2024relik}, and the inference patterns of trained models \cite{krishnan2024patterns}. We instead measure train--validation gaps and the effect of regularization across model families. Appendix~\ref{app:related} gives an extended review.

\section{Methodology}

\subsection{Preliminaries}

\subsubsection{Knowledge Graph}
Let the set of entities be denoted as $\mathcal{V}=\left\{v_1, \ldots, v_{|\mathcal{V}|}\right\}$, and the set of relations as $\mathcal{P}=\left\{\mathcal{R}_1, \ldots, \mathcal{R}_{|\mathcal{P}|}\right\}$. A knowledge graph is defined as a set of triples $\mathcal{T} \subseteq \mathcal{V} \times \mathcal{P} \times \mathcal{V}$, where each triple $( \mathcal{V}_i^h, \mathcal{R}_j^r, \mathcal{V}_k^t )$ corresponds to a head entity, a relation, and a tail entity, respectively. The knowledge graph (KG) is then represented as the tuple $\mathcal{G} = (\mathcal{V}, \mathcal{P}, \mathcal{T})$, comprising the set of entities, relations, and observed triples.

\subsubsection{Knowledge Graph Embedding and Link Prediction}
Given an observed knowledge graph $\mathcal{G} = (\mathcal{V}, \mathcal{P}, \mathcal{T})$, the task of link prediction is to infer the missing but plausible facts, denoted by $\mathcal{T}^\star$. The mainstream KGE method first assigns a vector embedding in real or complex space to each entity and relationship, and constructs a scoring function:
\begin{equation}
\phi_e: \mathcal{V} \rightarrow \mathbb{R}^{d_e} \text { or } \mathbb{C}^{d_e}, \quad \phi_\rho: \mathcal{P} \rightarrow \mathbb{R}^{d_\rho} \text { or } \mathbb{C}^{d_\rho},
\end{equation}
\begin{equation}
s: \mathcal{V} \times \mathcal{P} \times \mathcal{V} \longrightarrow \mathbb{R}, \quad\left(v_h, \rho_r, v_t\right) \mapsto s\left(v_h, \rho_r, v_t\right),
\end{equation}
which evaluates the plausibility of any candidate triple. For completion queries of the form $\left\langle v_i, \rho_j, ?\right\rangle$, the model substitutes the missing entity with each $v \in \mathcal{V}$ in turn and computes the corresponding score. Entities with the highest scores are top candidates for completing the triple.

\noindent\textbf{Notation.} For readability, throughout the paper we use $\mathcal{V}_i^h$, $v_h$, $h$, and $\mathbf{h}$ interchangeably to denote the head-entity embedding, and analogously $\mathcal{R}_j^r$, $\rho_r$, $r$, $\mathbf{r}$ for relations and $\mathcal{V}_k^t$, $v_t$, $t$, $\mathbf{t}$ for tail entities. The model-specific scoring functions introduced later keep the original notation of their source papers (e.g., $\mathbf{a}_h$, $\mathbf{b}_t$, $\mathbf{c}_r$ for CP); we use $\psi$ for the generic relation-conditioned interaction so as not to overload the embedding maps $\phi_e,\phi_\rho$ defined above.

\subsection{The SPR Regularizer Method}\label{sec:spr}

Existing KGE regularizers such as DURA \cite{zhang2020duality}, ER \cite{cao2022er}, and VIR \cite{xiao2024knowledge} have proven effective for tensor-decomposition models such as CP and ComplEx. However, their applicability across KGE model families has rarely been examined. SPR is designed for this broader setting: by sparsifying the regularization target, it focuses the penalty on the truly influential coordinates.

The basic idea of SPR is to sparsify the inputs to the regularization term so that only important components (i.e., components with large squared magnitudes) are penalized, while the many almost negligible components, which contribute little to the embedding norm, are left out of the penalty. This is achieved by sorting the (non-negative) squared terms and masking those whose cumulative sum is at most $\delta$, i.e., zeroing them in the regularization target. The masking applies \emph{only} to the regularization target, not to the embedding itself: the vectors consumed by the scoring function are never modified, and the low-energy coordinates remain in the embedding and continue to be shaped by the main task loss. What SPR changes is \emph{where the regularization budget is spent}. Instead of shrinking every coordinate uniformly as L2/Frobenius regularization does, it redirects the penalty onto the dominant, high-energy coordinates that most inflate the embedding norm and thereby the model capacity and the overfitting risk, while leaving the many near-zero coordinates penalty-free. These low-energy coordinates add little to the embedding norm and may still carry weak but useful signal, so penalizing them would barely reduce capacity while eroding such fine-grained features. A coordinate also cannot escape this control by staying small: because the mask is recomputed at every forward pass from the current values, once a low-energy coordinate grows large enough to matter for the norm, it is automatically returned to the penalized set. We use a fixed $\delta$ in this paper and leave an adaptive $\delta$, e.g., one depending on the gradient norm, to future work. SPR applies sparsity regularization to the head and tail entities and to their interaction terms with the relation; the relation embeddings are thus constrained through these interaction terms rather than by a separate penalty.
\subsubsection{Sorting and Selection}
We first define the binary mask $\mathcal{M}(\boldsymbol{x}) \in\{0,1\}^n$. For any non-negative vector $\boldsymbol{x} \in \mathbb{R}_{\geq 0}^n$ used as a regularization target (for example, the squared head-entity embeddings ${\mathcal{V}_i^h}^2$ of all triples in a mini-batch $B$, stacked into one vector of length $n=|B|D$), SPR selects the low-energy coordinates by their sorted indices. For models with complex-valued embeddings (e.g., ComplEx), the squaring is taken coordinate-wise on the modulus, i.e., $({\mathcal{V}_i^h}^2)_d=|{\mathcal{V}_i^h}_d|^2=\operatorname{Re}({\mathcal{V}_i^h}_d)^2+\operatorname{Im}({\mathcal{V}_i^h}_d)^2$, so that $\boldsymbol{x}$ is always real and non-negative. This modulus-based target is different from concatenating the real and imaginary parts before sorting, because the latter would double the length of the mask and can select different coordinates. Let $\pi$ be a permutation of $\{1,\ldots,n\}$ such that $\boldsymbol{x}_{\pi(1)} \leq \boldsymbol{x}_{\pi(2)} \leq \cdots \leq \boldsymbol{x}_{\pi(n)}$. Let $S$ be the largest integer such that:
\begin{equation}
\sum_{i=1}^{S} \boldsymbol{x}_{\pi(i)} \leq \delta.
\end{equation}
Concretely, the mask flags exactly the $S$ sorted indices: $\mathcal{M}(\boldsymbol{x})_{\pi(i)}=1$ for $i=1,\ldots,S$ and $\mathcal{M}(\boldsymbol{x})_{\pi(i)}=0$ for $i=S+1,\ldots,n$. Defining the mask by indices rather than by coordinate values is important when several entries of $\boldsymbol{x}$ are tied. The sparsified version of $\boldsymbol{x}$ is then given by:
\begin{equation}
\boldsymbol{x}_{\text {sparse }}=\boldsymbol{x} \odot(1-\mathcal{M}(\boldsymbol{x})),
\end{equation}
so that the small components (those for which $\mathcal{M}(\boldsymbol{x})_j=1$) are set to zero while the large ones are retained and penalized. The mask is computed from the current value of $\boldsymbol{x}$ and held fixed during back-propagation, and it is recomputed at every forward pass; hence the penalty gradient reaches only the retained coordinates, exactly as in a masked squared norm.

\textbf{Lemma 1 Sparsification Error Bound: }   \textit{Let $\boldsymbol{x}=\allowbreak\bigl(\boldsymbol{x}_1,\allowbreak \boldsymbol{x}_2,\allowbreak \ldots,\allowbreak \boldsymbol{x}_n\bigr)^{\top}\allowbreak \in \mathbb{R}_{\geq 0}^n$ be a nonnegative vector and suppose that the operator select-small$(\boldsymbol{x}, \delta)$ produces a binary mask $\mathcal{M} \in\{0,1\}^n$, such that the cumulative sum of the "dropped" entries is bounded by $\delta$, i.e., $\sum_{i: \mathcal{M}_i=1} \boldsymbol{x}_i \leq \delta$. Then the difference between the full sum and the sparsified sum is bounded as:}
\begin{equation}
\scalebox{0.85}{$\displaystyle
\left|\sum_{j=1}^n \boldsymbol{x}_j-\sum_{j=1}^n (\boldsymbol{x}_j)_{\text {sparse }}\right| \leq \delta.
$}
\end{equation}
The proof is given in Appendix~\ref{app:tech}.

\noindent\textbf{Remark.} Lemma~1 bounds the deviation of the sparsified penalty from the dense version of the \emph{same} regularization target by at most $\delta$; it does \emph{not} make SPR equivalent to plain squared-Frobenius (F2) regularization. The two differ both in the penalized quantities (SPR additionally constrains the entity--relation interaction terms) and in their gradients: SPR exerts no penalty gradient on the masked near-zero coordinates, whereas F2 shrinks every coordinate uniformly. Hence a small Lemma~1 gap is fully consistent with SPR and F2 behaving very differently during training.
\subsubsection{Sparsified Norms of Entities} Let $\mathbf{h}_B,\mathbf{r}_B,\mathbf{t}_B\in\mathbb{R}^{n}$ stack the head-entity, relation, and tail-entity embeddings of the triples in $B$ in the same order, and let $\boldsymbol{u}\in\{\mathbf{h}_B,\mathbf{t}_B\}$. Using the sparsification operator above, both entity terms share the compact form:
\begingroup
\small
\begin{equation}
\|\boldsymbol{u}\|_{\text{sparse}}^2
=\sum_{d=1}^{n} u_d^2\bigl(1-\mathcal{M}(\boldsymbol{u}^{\,2})_d\bigr).
\end{equation}
\endgroup

\subsubsection{Sparsified Interaction Terms} For vector-valued relations and the same $\boldsymbol{u}$, both the head- and tail-entity interactions with the relation share the form:
\begingroup
\small
\begin{equation}
\|\boldsymbol{u}\odot\mathbf{r}_B\|_{\text{sparse}}^2
=\sum_{d=1}^{n} u_d^2 r_{B,d}^2
\bigl(1-\mathcal{M}(\boldsymbol{u}^{\,2}\odot\mathbf{r}_B^{\,2})_d\bigr).
\end{equation}
\endgroup

For matrix-relation models such as RESCAL, the relation is not a length-$D$ vector but a matrix $M_r \in \mathbb{R}^{D \times D}$. In this case, SPR keeps the same entity terms and defines the relation-conditioned interaction terms on the transformed entity representations that enter the bilinear score, i.e., on $M_r^{\top}\mathcal{V}_i^h$ and $M_r\mathcal{V}_k^t$ stacked over $B$.
Combining the above sparse terms, the SPR regularization term of a mini-batch with vector-valued relation embeddings is:
\begingroup
\small
\begin{equation}
\begin{aligned}
\operatorname{SPR}\bigl(\mathbf{h}_B,\; \mathbf{r}_B,\; \mathbf{t}_B\bigr)
&= \left\|\mathbf{h}_B\right\|_{\text{sparse}}^2
+ \left\|\mathbf{t}_B\right\|_{\text{sparse}}^2
\\ &\quad + \left\|\mathbf{h}_B \odot \mathbf{r}_B\right\|_{\text{sparse}}^2
+ \left\|\mathbf{t}_B \odot \mathbf{r}_B\right\|_{\text{sparse}}^2.
\end{aligned}
\end{equation}
\endgroup
When SPR is applied to RESCAL, the two element-wise interaction terms in this expression are replaced by the matrix-based interaction terms defined above. With $\delta=0$, all masks are empty and Eq.~(8) reduces to the unweighted DURA penalty~\cite{zhang2020duality}.
The overall regularization loss of the mini-batch is:
\begingroup
\small
\begin{equation}
\Omega_{\text {SPR }}=\frac{1}{|B|}\operatorname{SPR}\bigl(\mathbf{h}_B,\;\mathbf{r}_B,\;\mathbf{t}_B\bigr).
\end{equation}
\endgroup
The complete pseudocode is given in Algorithm~\ref{alg:spr} (Appendix~\ref{app:spr-pseudocode}).

For a given triple, many translation-based (TB) and tensor-\allowbreak decomposition-based (TDB) KGE models can be written in a unified schematic form:
\begin{equation}
f_{\mathcal{R}_j^r}\left(\mathcal{V}_i^h, \mathcal{V}_k^t\right) = \psi\left(\mathcal{V}_i^h , \mathcal{R}_j^r, \mathcal{V}_k^t\right) ,
\end{equation}
where $\psi$ first lets the relation $\mathcal{R}_j^r$ act on the entities and then outputs a plausibility score. For distance-based TB models, we use the negative distance as the score (e.g., $-\|\mathcal{V}_i^h+\mathcal{R}_j^r-\mathcal{V}_k^t\|$ for TransE), whereas TDB models use a (bi)linear similarity score (e.g., ${\mathcal{V}_i^h}^{\top} M_{\mathcal{R}_j^r}\mathcal{V}_k^t$ of RESCAL or the Hermitian product of ComplEx). Models with richer geometry or neural message passing, such as GIE and CompGCN~\cite{vashishth2019composition}, instantiate $\psi$ differently, yet SPR remains applicable because it constrains the entity embeddings and their interaction terms with the relation, which all these models share. Each backbone keeps the training loss of its original implementation; for CP, ComplEx, and RESCAL, we follow \cite{lacroix2018canonical,zhang2020duality} and use the cross-entropy over all candidate entities:
\begin{equation}
\mathcal{L}_p=-\sum_{(h,r,t)\in\mathcal{G}}\log\frac{\exp\bigl(f_r(h,t)\bigr)}{\sum_{t'\in\mathcal{V}}\exp\bigl(f_r(h,t')\bigr)},
\end{equation}
where $\mathcal{G}$ also contains the reciprocal triple $(t,r^{-1},h)$ of every training triple. With a regularizer, the objective becomes:
\begin{equation}
\min \; \mathcal{L}_p+\lambda \Omega,
\end{equation}
where $\Omega(\cdot)$ is the regularizer and $\lambda$ is the regularization coefficient.
\subsubsection{Score Function} CP \cite{hitchcock1927expression} is a classic tensor-factorization model that represents the knowledge graph tensor as a sum of rank-one tensors. It is fully expressive and parameter-efficient. Its score function is:
\begin{equation}
f_{\mathrm{CP}}(h, r, t)=\sum_{k=1}^d a_{h, k} b_{t, k} c_{r, k},
\end{equation}
where $a_{h, k}$ is the $k^{\text {th }}$ component of the subject side embedding $\mathbf{a}_h \in \mathbb{R}^d$; $b_{t, k}$ is the $k^{\text {th }}$ component of the object side embedding $\mathbf{b}_t \in \mathbb{R}^d$;
$c_{r, k}$ is the $k^{\text {th }}$ component of the relation vector $\mathbf{c}_r \in \mathbb{R}^d$; and $d$ is the embedding dimensionality.

ComplEx \cite{trouillon2016complex} extends DistMult\cite{yang2014embedding} into the complex plane so it can natively model asymmetric relations while keeping $O(d)$ parameters per relation. The real part of a trilinear Hermitian product is used as the plausibility score:
\begin{equation}
f_{\mathrm{ComplEx}}(h, r, t)=\operatorname{Re}\left(\sum_{k=1}^d h_k r_k \overline{t_k}\right) ,
\end{equation}
where $h_k, r_k, t_k \in \mathbb{C}$ are the $k^{\text {th }}$ complex components of entity and relation embeddings; $\overline{t_k}$ is the complex conjugate of $t_k$; $\operatorname{Re}(\cdot)$ extracts the real part, turning a complex score into a real scalar.

RESCAL \cite{nickel2011three} learns one dense matrix per relation so that every pair of latent features can interact:
\begin{equation}
f_{\mathrm{RESCAL}}(h, r, t)=\mathbf{h}^{\top} M_r \mathbf{t} ,
\end{equation}
where $\mathbf{h}, \mathbf{t} \in \mathbb{R}^d$ are entity vectors shared across relations; $M_r \in \mathbb{R}^{d \times d}$ is the relation-specific matrix; and $\mathbf{h}^{\top} M_r \mathbf{t}$ is a bilinear form that produces the triple score.

GIE \cite{cao2022geometry}, a typical translation-based model, embeds every entity in three curved spaces (Euclidean $\mathcal{E}$, hyperbolic $\mathcal{H}$, hyperspherical $\mathcal{S}$) and scores a triple by interacting those geometries via cross-space geodesic distances:
\begin{equation}
\begin{aligned}
f_{\mathrm{GIE}}(h, r, t)
={}& -\Bigl(d_c\bigl(\operatorname{Inter}(E_h, H_h, S_h), \mathbf{t}\bigr)\\
&\quad + d_c\bigl(\operatorname{Inter}(E_t, H_t, S_t), \mathbf{h}\bigr)\Bigr) + b
\end{aligned}
\end{equation}
where the relation $r$ enters the score through the relation-specific geometric transformations applied to the entities: $E_h, H_h, S_h$ denote the head entity after it has been transformed by $r$ in the Euclidean, hyperbolic, and hyperspherical manifolds, respectively (in Euclidean space this is simply $E_h = r\,h$, and the hyperbolic and hyperspherical counterparts are obtained analogously through the relation-specific M\"obius transformation of $r$); symmetrically, $E_t, H_t, S_t$ denote the tail entity transformed by the reverse relation $r'$; $\operatorname{Inter}(\cdot)$ is the learned attention-weighted fusion of the three points; $d_c$ is the manifold geodesic distance with curvature $c$; and $b$ is a trainable bias. Hence the relation $r$ (together with its reverse $r'$) directly parameterizes the transformations of both entities, so that different relations yield different scores for the same entity pair $(h,t)$, as required for relation-conditioned link prediction.
\begin{table*}[!t]
\centering
\caption{Main link prediction results on WN18RR, FB15K-237, and YAGO3-10; all models were run on the same machine. Within each dataset block, bold and underlined values are the best and second-best in each column (ties are marked alike).}
\label{tab:kgc-main-dataset}
\footnotesize
\setlength{\tabcolsep}{2.4pt}
\renewcommand{\arraystretch}{0.94}
\begin{tabularx}{\textwidth}{@{}l@{\hspace{8pt}}l*{16}{>{\centering\arraybackslash}X}@{}}
\toprule
\multirow{2}{*}{Dataset} & \multirow{2}{*}{Reg.} & \multicolumn{4}{c}{CP} & \multicolumn{4}{c}{ComplEx} & \multicolumn{4}{c}{GIE} & \multicolumn{4}{c}{RESCAL} \\
\cmidrule(lr){3-6} \cmidrule(lr){7-10} \cmidrule(lr){11-14} \cmidrule(l){15-18}
& & MRR & H@1 & H@3 & H@10 & MRR & H@1 & H@3 & H@10 & MRR & H@1 & H@3 & H@10 & MRR & H@1 & H@3 & H@10 \\
\midrule
\multirow{6}{*}{WN18RR} & No-Reg & 0.438 & 0.416 & 0.465 & 0.485 & 0.457 & 0.426 & 0.471 & 0.518 & 0.464 & 0.434 & 0.479 & 0.529 & 0.444 & 0.419 & 0.453 & 0.491 \\
 & F2 & 0.449 & 0.421 & 0.474 & 0.506 & 0.468 & 0.431 & 0.479 & 0.535 & 0.470 & 0.439 & 0.489 & 0.553 & 0.451 & 0.424 & 0.462 & 0.511 \\
 & N3 & 0.469 & 0.432 & 0.485 & 0.541 & \underline{0.488} & 0.445 & \underline{0.505} & \underline{0.570} & \underline{0.491} & \underline{0.452} & \underline{0.507} & \underline{0.576} & 0.475 & 0.437 & 0.479 & 0.556 \\
 & DURA & 0.471 & 0.433 & 0.488 & 0.545 & \underline{0.488} & \underline{0.447} & 0.502 & 0.568 & 0.488 & 0.449 & 0.503 & 0.571 & \underline{0.495} & \underline{0.454} & \underline{0.511} & \underline{0.576} \\
 & ER & \underline{0.475} & \underline{0.436} & \underline{0.489} & \underline{0.549} & 0.486 & 0.444 & 0.502 & 0.566 & 0.490 & 0.450 & 0.505 & 0.574 & 0.493 & 0.452 & 0.505 & 0.575 \\
\rowcolor{gray!15}\cellcolor{white} & SPR & \textbf{0.479} & \textbf{0.440} & \textbf{0.491} & \textbf{0.555} & \textbf{0.491} & \textbf{0.451} & \textbf{0.509} & \textbf{0.572} & \textbf{0.497} & \textbf{0.455} & \textbf{0.514} & \textbf{0.581} & \textbf{0.499} & \textbf{0.456} & \textbf{0.518} & \textbf{0.581} \\
\midrule
\multirow{6}{*}{FB15K-237} & No-Reg & 0.327 & 0.239 & 0.359 & 0.502 & 0.350 & 0.260 & 0.384 & 0.532 & 0.341 & 0.254 & 0.373 & 0.520 & 0.350 & 0.262 & 0.382 & 0.525 \\
 & F2 & 0.332 & 0.250 & 0.369 & 0.517 & 0.352 & 0.261 & 0.387 & 0.537 & 0.349 & 0.260 & 0.387 & 0.539 & 0.355 & 0.268 & 0.391 & 0.538 \\
 & N3 & 0.354 & 0.260 & 0.389 & 0.543 & 0.360 & 0.268 & 0.399 & 0.551 & \underline{0.364} & \underline{0.269} & \underline{0.400} & \underline{0.555} & 0.361 & 0.270 & 0.397 & 0.547 \\
 & DURA & \underline{0.360} & \underline{0.266} & \underline{0.395} & \underline{0.550} & 0.365 & 0.270 & 0.403 & 0.555 & 0.355 & 0.262 & 0.395 & 0.547 & \underline{0.365} & \underline{0.273} & \underline{0.403} & 0.546 \\
 & ER & 0.358 & 0.263 & 0.394 & 0.549 & \underline{0.369} & \underline{0.275} & \underline{0.408} & \underline{0.561} & 0.351 & 0.259 & 0.388 & 0.541 & 0.363 & 0.270 & 0.401 & \underline{0.549} \\
\rowcolor{gray!15}\cellcolor{white} & SPR & \textbf{0.366} & \textbf{0.272} & \textbf{0.404} & \textbf{0.558} & \textbf{0.371} & \textbf{0.277} & \textbf{0.409} & \textbf{0.566} & \textbf{0.367} & \textbf{0.272} & \textbf{0.402} & \textbf{0.558} & \textbf{0.367} & \textbf{0.276} & \textbf{0.407} & \textbf{0.551} \\
\midrule
\multirow{6}{*}{YAGO3-10} & No-Reg & 0.567 & 0.495 & 0.609 & 0.695 & 0.569 & 0.495 & 0.614 & 0.703 & 0.574 & 0.503 & 0.617 & 0.699 & 0.551 & 0.471 & 0.598 & 0.695 \\
 & F2 & 0.571 & 0.499 & 0.617 & 0.699 & 0.573 & 0.496 & 0.616 & 0.710 & 0.574 & 0.504 & 0.618 & 0.701 & 0.554 & 0.473 & 0.602 & 0.699 \\
 & N3 & 0.573 & 0.499 & 0.618 & 0.705 & 0.574 & 0.498 & 0.618 & 0.710 & 0.579 & 0.508 & 0.623 & 0.710 & 0.556 & 0.479 & 0.604 & 0.703 \\
 & DURA & \underline{0.577} & \underline{0.504} & \underline{0.623} & \underline{0.709} & \underline{0.581} & 0.506 & 0.624 & \underline{0.716} & \underline{0.586} & 0.515 & 0.623 & 0.714 & 0.563 & 0.486 & 0.608 & 0.704 \\
 & ER & 0.576 & 0.502 & 0.620 & 0.706 & \textbf{0.584} & \underline{0.512} & \underline{0.626} & 0.712 & 0.585 & \underline{0.517} & \underline{0.626} & \textbf{0.721} & \underline{0.569} & \underline{0.502} & \underline{0.611} & \underline{0.709} \\
\rowcolor{gray!15}\cellcolor{white} & SPR & \textbf{0.582} & \textbf{0.509} & \textbf{0.627} & \textbf{0.713} & \textbf{0.584} & \textbf{0.514} & \textbf{0.628} & \textbf{0.719} & \textbf{0.593} & \textbf{0.522} & \textbf{0.629} & \underline{0.720} & \textbf{0.575} & \textbf{0.511} & \textbf{0.619} & \textbf{0.713} \\
\bottomrule
\end{tabularx}
\end{table*}

\section{Experiment}\label{sec:experiment}
Our experiments answer five research questions (RQs):
\begingroup
\setlength{\parskip}{0pt}
\par \textbf{RQ1 (Necessity):} Do existing KGE models generally overfit, and is regularization necessary to mitigate this overfitting?\par
\par \textbf{RQ2 (Performance):} Under the same training protocol (model, optimizer, loss, and training budget), does SPR achieve higher MRR and Hits@K than common regularizers and keep this advantage across benchmarks?\par
\par \textbf{RQ3 (Applicability):} Does SPR consistently improve KGE models of different architectures, including GNN-based, temporal, and multimodal ones?\par
\par \textbf{RQ4 (Sensitivity):} How sensitive is SPR to its hyperparameters, is there a stable range of good values, and what happens beyond it?\par
\par \textbf{RQ5 (Visualization):} What qualitative differences appear in two-dimensional projections of entity embeddings learned under different regularizers?\par
\endgroup

\subsection{Experiment Setting}\label{sec:setting}
\subsubsection{Datasets}
We use eight datasets: the static benchmarks WN18RR \cite{dettmers2018convolutional}, FB15K-237 \cite{toutanova2015observed}, YAGO3-10 \cite{mahdisoltani2014yago3}, Kinship \cite{kemp2006learning}, and UMLS \cite{mccray2003upper}; the temporal benchmarks ICEWS14 and ICEWS05-15 \cite{garcia2018learning}; and the multimodal benchmark WN9-IMG \cite{xie2016image}. Appendix~\ref{app:datasets} describes them and reports their statistics.

\subsubsection{Protocol and Metrics}
To ensure the same training protocol, all regularization variants of the same backbone on the same dataset use the same optimizer, loss, batch size, embedding dimension, training budget, and early-stopping criterion; only the regularization term and its coefficient-related hyperparameters are changed and selected on the validation set. We report the Mean Reciprocal Rank (MRR), Hits@$N$ ($N\in\{1,3,10\}$), and the Mean Rank (MR). Appendix~\ref{app:setup} gives the implementation details, hyperparameter ranges, and metric definitions.

\subsection{Effects of Regularization on KGE Training Curves (RQ1)}
\begin{figure*}[t]
    \centering
    \includegraphics[width=0.8\linewidth]{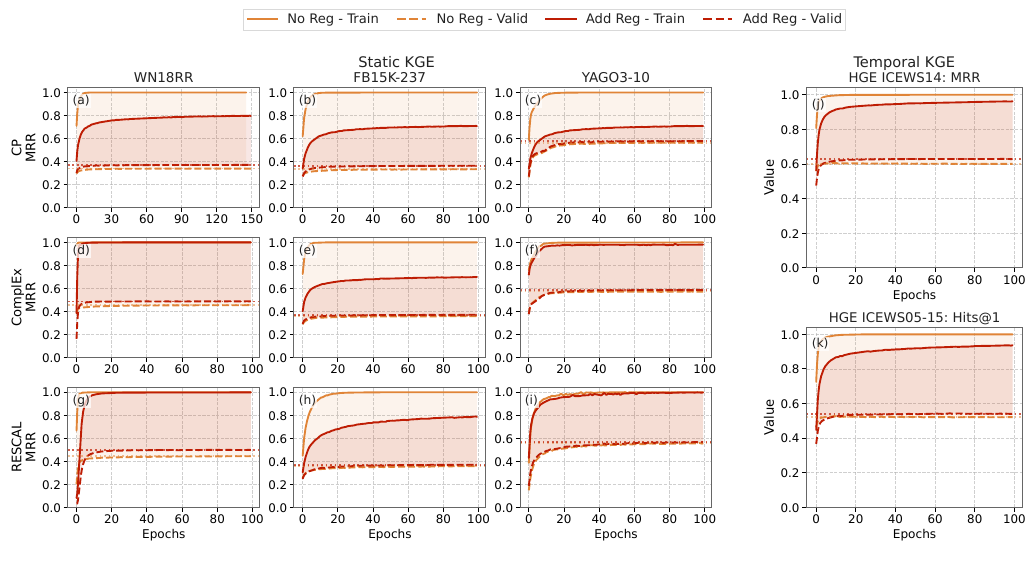}
    \caption{Training curves with and without regularization across static and temporal KGE models. Panels (a)--(i) report MRR for CP, ComplEx, and RESCAL (rows) on WN18RR, FB15K-237, and YAGO3-10 (columns); panels (j) and (k) report HGE on ICEWS14 (MRR) and ICEWS05-15 (Hits@1), respectively. Orange and red denote No Reg and Add Reg; solid and dashed curves denote training and validation, dotted horizontal lines mark the final validation performance after convergence, and shading marks the train--validation gap. Regularization consistently narrows the gap and improves validation performance.}
    \label{fig-A-overfit-curves}
\end{figure*}

Figure~\ref{fig-A-overfit-curves} tracks how training and validation performance evolve with and without regularization on static and temporal KGE models. Without regularization, every backbone reaches a near-perfect training score within a few epochs, while its validation score saturates at a much lower level and stays there for the rest of training. A model that behaves this way has memorized the observed triples instead of learning patterns that transfer to unseen facts. Existing KGE regularizers such as N3~\cite{lacroix2018canonical}, DURA~\cite{zhang2020duality}, ER~\cite{cao2022er}, and VIR~\cite{xiao2024knowledge} were designed and validated almost exclusively on tensor-decomposition models, which left open whether overfitting also affects other families and whether regularization helps there. Read together with the translation-based and GNN-based results reported elsewhere in this paper, these curves settle that question: the problem is general rather than specific to one model family.

Regularization improves the picture in every panel. The regularized validation curve ends above its unregularized counterpart and the gap to the training curve narrows, at the cost of slower convergence and a slightly lower training peak. That is an acceptable price, since only performance on unseen facts matters in practice. HGE follows the same pattern in the temporal setting: although it is designed to capture time-evolving relational patterns, without a penalty term it also saturates on the training data and generalizes poorly. Overfitting is thus inherent to representation learning on knowledge graphs rather than a quirk of the static setting, and regularization is a necessary component in both.

\subsection{Main Performance (RQ1 \& RQ2)}

Table~\ref{tab:kgc-main-dataset} reports the results of CP, ComplEx, GIE, and RESCAL with different regularizers on WN18RR, FB15K-237, and YAGO3-10. Regularization improves every evaluated KGE model, and SPR performs best overall: it obtains the best result in 47 of the 48 backbone--metric combinations, the only exception being the Hits@10 of GIE on YAGO3-10, where it trails ER by 0.001.

KGE models without regularization tend to overfit: they perform well on the training data but poorly on the validation and test data. Accordingly, the unregularized models score lowest on all datasets, especially on ranking metrics such as Hits@1 and Hits@3, which indicates that they overfit the training data or fail to learn high-quality embeddings. After adding regularization, performance improves. Taking CP on WN18RR as an example, the MRR of CP-SPR is 0.479, an improvement of about 9.4\% over the base model. Similar trends hold for other models and datasets; for example, on FB15K-237, ComplEx-SPR raises the MRR of the base ComplEx model from 0.350 to 0.371. These results support \textbf{RQ1 and RQ2}.

\begin{table}[t]
\centering
\caption{Results on UMLS and Kinship. Bold and underlined values are the best and second-best in each column.}
\label{tab:kgc-small-dataset}
\footnotesize
\setlength{\tabcolsep}{3.5pt}
\begin{tabularx}{\columnwidth}{@{}l*{8}{>{\centering\arraybackslash}X}@{}}
\toprule
\multirow{3}{*}{Reg.} & \multicolumn{4}{c}{CP} & \multicolumn{4}{c}{ComplEx} \\
\cmidrule(lr){2-5} \cmidrule(l){6-9}
& \multicolumn{2}{c}{UMLS} & \multicolumn{2}{c}{Kinship} & \multicolumn{2}{c}{UMLS} & \multicolumn{2}{c}{Kinship} \\
\cmidrule(lr){2-3} \cmidrule(lr){4-5} \cmidrule(lr){6-7} \cmidrule(l){8-9}
& MRR & H@1 & MRR & H@1 & MRR & H@1 & MRR & H@1 \\
\midrule
No-Reg & 0.868 & 0.780 & 0.872 & 0.784 & 0.897 & 0.822 & 0.875 & 0.783 \\
N3     & 0.879 & 0.782 & 0.885 & 0.819 & 0.907 & 0.843 & 0.883 & 0.786 \\
DURA   & 0.881 & 0.788 & 0.885 & 0.792 & 0.904 & 0.841 & \textbf{0.888} & \underline{0.796} \\
ER     & \underline{0.883} & \underline{0.791} & \underline{0.888} & \underline{0.820} & \textbf{0.910} & \underline{0.844} & \underline{0.887} & 0.795 \\
\rowcolor{gray!15}SPR    & \textbf{0.885} & \textbf{0.792} & \textbf{0.889} & \textbf{0.821} & \underline{0.909} & \textbf{0.847} & \underline{0.887} & \textbf{0.820} \\
\bottomrule
\end{tabularx}
\end{table}

To verify SPR beyond large-scale datasets such as YAGO3-10, we further evaluate it on the two small datasets UMLS and Kinship, with results in Table~\ref{tab:kgc-small-dataset}. SPR is best on all four metrics for CP and on both Hits@1 metrics for ComplEx, and its MRR is within 0.001 of the best in the remaining two cases. This confirms that selectively regularizing only the important components of the embedding space is effective, and that SPR alleviates overfitting and improves the prediction ability of KGE models on both large and small graphs.

\subsection{Overfitting in Multimodal Knowledge Graph Embedding Models (RQ1 \& RQ3)}\label{sec:mmkge}
Multimodal knowledge graph embedding (MMKGE) models enrich structural triples with additional modality information, such as images or text \cite{liang2024survey}. Such side information can provide more semantic clues for link prediction, but it does not automatically remove the risk of memorizing the training graph. To examine whether the overfitting phenomenon still exists in this broader setting, we further evaluate a representative multimodal KGE backbone, OTKGE~\cite{NEURIPS2022_ffdb280e}, on the WN9-IMG benchmark~\cite{xie2016image}.

\begin{figure*}[t]
    \centering
    \includegraphics[width=0.8\linewidth]{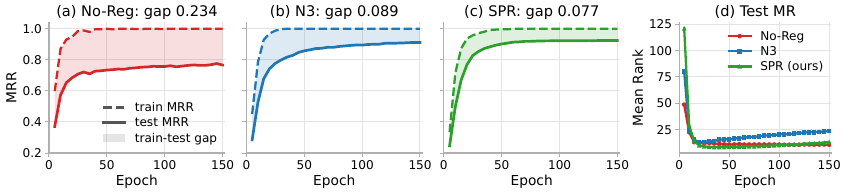}
    \caption{Overfitting analysis of OTKGE on WN9-IMG. (a)--(c) Train (dashed) and test (solid) MRR under No-Reg, N3, and SPR; the shaded area is the train--test gap, whose final value is given in each panel title. (d) Test MR of the three settings.}
    \label{fig:mmkge-otkge-overfit}
\end{figure*}

As shown in Figure~\ref{fig:mmkge-otkge-overfit}, the unregularized OTKGE model still exhibits a clear overfitting pattern: its training MRR rapidly approaches 1.0, whereas its test MRR converges to a much lower level, resulting in a large final train--test gap of 0.234. This indicates that multimodal signals alone are insufficient to guarantee good generalization, and MMKGE models still require proper regularization. After introducing regularization, both N3 and SPR reduce the generalization gap and improve the test MRR. SPR reaches the highest test MRR of the three settings, and compared with N3 it also has a smaller final train--test gap (0.077 vs.\ 0.089) and a lower test MR. These observations extend our answer to \textbf{RQ1}: overfitting is not limited to conventional KGE models but also appears in multimodal KGE models. They also support \textbf{RQ3}, showing that SPR remains effective when the backbone incorporates multimodal information.

\subsection{Applicability across Model Families (RQ3)}

\begin{figure*}[t]
    \centering
    \includegraphics[width=0.9\linewidth,trim=0 7 0 7,clip]{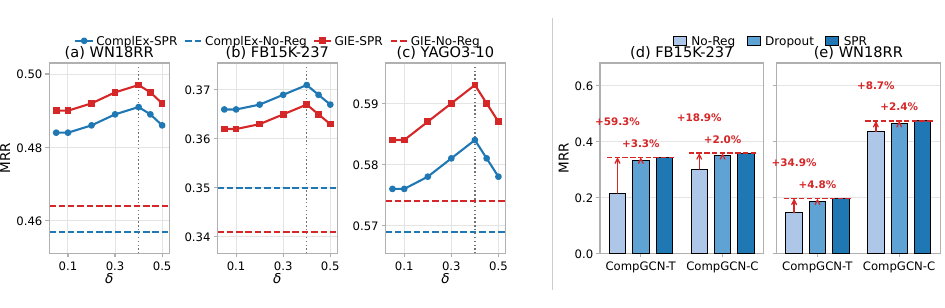}
    \caption{MRR under (a)--(c) different SPR thresholds $\delta$, with no-regularization baselines (dashed) and $\delta=0.4$ (dotted), and (d)--(e) No-Reg, Dropout, and SPR for CompGCN-TransE (CompGCN-T) and CompGCN-ConvE (CompGCN-C); red arrows give the relative gains of SPR.}
    \label{fig:delta-gnn}
\end{figure*}

Besides OTKGE (Section~\ref{sec:mmkge}), we evaluate SPR with four types of backbones: the translation-based GIE, the tensor-\allowbreak decomposition-based ComplEx, the GNN-based CompGCN~\cite{vashishth2019composition}, and the temporal model HGE. Figure~\ref{fig:delta-gnn}(d)--(e) reports the CompGCN results. The three bars represent separate training settings. Both Dropout and SPR improve MRR over No-Reg, and SPR reaches the highest value in all four groups: 0.344 and 0.359 on FB15K-237 and 0.197 and 0.474 on WN18RR for CompGCN-T and CompGCN-C, respectively. These values correspond to gains of 3.3\%, 2.0\%, 4.8\%, and 2.4\% over Dropout. SPR regularizes entity embeddings and entity--relation interaction terms without assuming a particular scoring function, which allows it to be applied to message-passing KGE models.

We further test regularization in the temporal setting \cite{wang2023survey,cai2022temporal}, training HGE \cite{pan2024hge} on ICEWS14 and ICEWS05-15 \cite{garcia2018learning}. As Table~\ref{tab:icews-results} shows, SPR improves MRR by about 4.2\% and 5.4\% over the unregularized baseline and Hits@10 by about 6.9\% and 5.9\%, with the remaining metrics gaining between 2.7\% and 4.6\%. It also outperforms N3 on every metric of both datasets, so a time-aware backbone benefits from the same selective penalty.

\begin{table}[t]
\centering
\caption{Performance of HGE-based temporal KGE on ICEWS14 and ICEWS05-15 under different regularization strategies (No-Reg, N3, SPR).}
\label{tab:icews-results}
\footnotesize
\setlength{\tabcolsep}{3.2pt}
\begin{adjustbox}{max width=\columnwidth}
\begin{tabular}{@{}lcccccccc@{}}
\toprule
\multirow{2}{*}{Model} & \multicolumn{4}{c}{ICEWS14} & \multicolumn{4}{c}{ICEWS05-15} \\
\cmidrule(lr){2-5} \cmidrule(l){6-9}
& MRR & H@1 & H@3 & H@10 & MRR & H@1 & H@3 & H@10 \\
\midrule
HGE+No-Reg & 0.596 & 0.514 & 0.648 & 0.739 & 0.653 & 0.579 & 0.704 & 0.785 \\
HGE+N3 & 0.617 & 0.526 & 0.673 & 0.783 & 0.684 & 0.599 & 0.733 & 0.829 \\
\rowcolor{gray!15}HGE+SPR & \textbf{0.621} & \textbf{0.528} & \textbf{0.678} & \textbf{0.790} & \textbf{0.688} & \textbf{0.602} & \textbf{0.736} & \textbf{0.831} \\
\bottomrule
\end{tabular}
\end{adjustbox}
\end{table}

\begin{figure*}[t]
    \centering
    \includegraphics[width=0.9\linewidth,trim=0 5 0 3,clip]{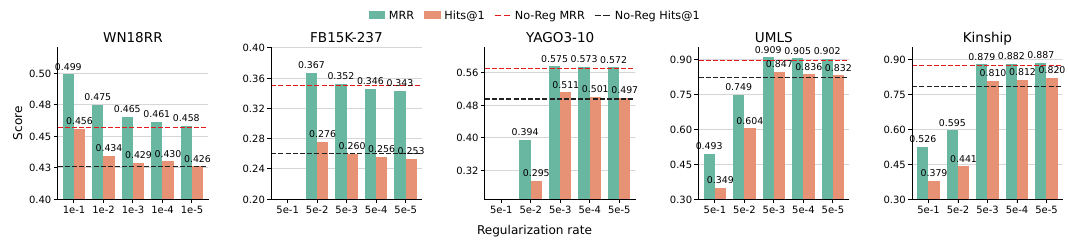}
    \caption{Sensitivity of ComplEx to the SPR regularization rate on five benchmarks. Green and orange bars report MRR and Hits@1, respectively; red and black dashed lines mark the corresponding no-regularization baselines.}
    \label{fig:reg-rate-sensitivity}
\end{figure*}

\subsection{Sensitivity Experiments (RQ4)}

We examine how the SPR regularization rate influences the MRR and Hits@1 of ComplEx on each dataset. Regularization trades off overfitting against underfitting: a rate that is too small leaves the model overfitting, and one that is too large makes it underfit. Performance is therefore usually non-monotonic in the rate, first increasing and then decreasing.

Figure~\ref{fig:reg-rate-sensitivity} summarizes the sensitivity results across five datasets, arranged from left to right as WN18RR, FB15K-237, YAGO3-10, UMLS, and Kinship.
The results confirm this trade-off: performance varies non-monotonically with the regularization rate, and both excessively high and excessively low rates can degrade performance, even below that of no regularization. On FB15K-237 and YAGO3-10, an overly large rate imposes such a severe penalty that the model fails to learn meaningful semantic information about entities. This leads to embedding collapse, with metric values dropping below 0.1 (these bars vanish from the corresponding columns in the figure).

With a properly tuned rate, SPR improves both MRR and Hits@1 over the unregularized baseline on all five datasets, showing that it mitigates overfitting and improves generalization.

The improvement varies across datasets: it is more pronounced on WN18RR and FB15K-237 and relatively small on YAGO3-10. This may reflect dataset characteristics such as scale, sparsity, and relation complexity, analyzed in Section~\ref{sec:dataset-composition}.

To quantify the impact of the threshold $\delta$, we conducted sensitivity experiments on three datasets with two backbones, ComplEx and GIE. We varied $\delta$ over the set $\{0.05, 0.1, 0.2, 0.3, 0.4, 0.45, 0.5\}$, and compared each SPR curve with its no-regularization baseline in Figure~\ref{fig:delta-gnn}(a)--(c). The results exhibit a clear non-monotonic trend: performance improves as $\delta$ increases from small values, reaches its best value around $\delta=0.4$, and then decreases when the threshold becomes larger. Across all three datasets and both backbones, every SPR curve stays above the corresponding no-regularization baseline over the whole range of $\delta$, so SPR is robust to the choice of the threshold. All six curves peak at $\delta=0.4$, making it a good default.
\subsection{Embedding Visualization Experiment (RQ5)}
\begin{figure}[t]
    \centering
    \setlength{\tabcolsep}{0pt}
    \footnotesize
    \begin{tabular*}{\linewidth}{@{\extracolsep{\fill}}cccc@{}}
    \includegraphics[width=0.242\linewidth,trim=6.6 41.0 970.5 20.2,clip]{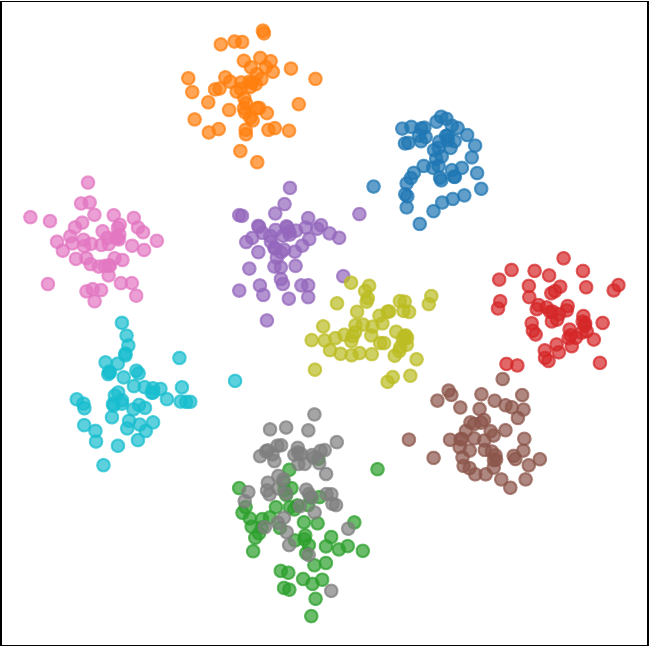} &
    \includegraphics[width=0.242\linewidth,trim=327.9 41.0 649.2 20.2,clip]{figure/regulation-visual2.pdf} &
    \includegraphics[width=0.242\linewidth,trim=649.2 41.0 327.9 20.2,clip]{figure/regulation-visual2.pdf} &
    \includegraphics[width=0.242\linewidth,trim=970.5 41.0 6.6 20.2,clip]{figure/regulation-visual2.pdf} \\[1pt]
    (a) SPR & (b) DURA & (c) ER & (d) No Reg
    \end{tabular*}
    \caption{ComplEx entity embeddings on WN18RR under (a) SPR, (b) DURA, (c) ER, and (d) no regularization, projected by t-SNE and colored by K-Means clusters ($K{=}10$); colors are not matched across panels.}
    \label{fig:vis}
\end{figure}

We visualize the entity embeddings learned by ComplEx with SPR, DURA, ER, and no regularization on WN18RR. For each setting, K-Means with $K=10$ is applied to the learned embeddings, and the embeddings are then projected to two dimensions using t-SNE. In Figure~\ref{fig:vis}, SPR yields compact clusters with clear boundaries for most groups, whereas without regularization more clusters overlap, such as the orange and purple groups and the red and brown groups in panel (d). Since K-Means and t-SNE are run independently for each setting, this comparison is qualitative.

\subsection{Dataset Scale and Regularization Gains}\label{sec:dataset-composition}
The relative improvements of the models on the three main datasets are not uniform, and the gap is largely a matter of scale. The training set grows from WN18RR (about 87K triples over 11 relations) to FB15K-237 (about 272K over 237 relations) to YAGO3-10 (about 1.08M over 37 relations), so the average number of training triples per relation is about 7.9K, 1.1K, and 29K, respectively.

On YAGO3-10 (the largest training set), the relative Hits@1 improvement brought by SPR over the unregularized base model is generally the smallest: for CP, ComplEx, and GIE it stays at roughly 2\%--4\% (about 2.8\%, 3.8\%, and 3.8\%, respectively), the only clear exception being RESCAL (about 8.5\%), whose unregularized baseline on YAGO3-10 is itself unusually weak. By contrast, the two smaller datasets, WN18RR and FB15K-237, generally obtain larger and broadly comparable gains (mostly in the 5\%--14\% range), with the single largest improvement observed for CP on FB15K-237 (about 13.8\%). These observations suggest that larger training sets may expose models to more diverse patterns, reducing the tendency to memorize noise or atypical samples. Regularization may therefore provide smaller gains on larger datasets and play a more important role on datasets with fewer training triples.

\section{Conclusion}
In this paper, we rethink regularization in KGE models. Our first contribution is an extensive empirical study: unlike prior work validated mainly on tensor-decomposition models, it spans translation-based, tensor-\allowbreak decomposition-based, GNN-based, temporal, and multimodal models, and shows that overfitting is a common yet frequently overlooked problem across all of them, while proper regularization greatly improves their generalization and performance relative to their unregularized baselines.

Our second contribution is SPR, a novel yet simple regularization method for KGE. SPR mitigates overfitting by concentrating the penalty on the dominant, high-energy components that most inflate the embedding norm, while leaving the low-energy components penalty-free. Extensive experiments on eight datasets and diverse KGE models show that SPR achieves the best results in nearly all settings compared with existing regularization methods.
\label{sec:main-end}

\bibliographystyle{plainnat}
\bibliography{references}

\appendix
\section{Technical Appendices and Proofs}
\label{app:tech}

\textbf{Proof of Lemma~1.} Since $\boldsymbol{x}_i\,(1-\mathcal{M}_i)$ retains exactly the unmasked entries,
\begin{equation*}
\sum_{i=1}^n \boldsymbol{x}_i-\sum_{i=1}^n \boldsymbol{x}_i\left(1-\mathcal{M}_i\right)=\sum_{i:\mathcal{M}_i=1} \boldsymbol{x}_i \in [0,\delta],
\end{equation*}
because every $\boldsymbol{x}_i\ge 0$ and the dropped mass is at most $\delta$ by construction of select-small$(\boldsymbol{x},\delta)$. Hence the absolute difference is at most $\delta$. $\hfill\square$

\subsection{SPR vs.\ Dropout Theory}\label{app:capacity}

Here $\mathbf{x}\in\mathbb{R}^D$ is an embedding vector in a mini-batch. Dropout \cite{srivastava2014dropout} zeroes coordinates with an i.i.d.\ mask $\mathbf{m}\sim\operatorname{Bernoulli}(1-p)^D$ that is \emph{independent} of their values. We analyze its non-inverted form (no division by $1-p$ during training) with the same squared-norm penalty applied to the vector that enters the loss, whose expectation is $\mathbb{E}_{\mathbf{m}}\|\mathbf{x}\odot\mathbf{m}\|_2^{2}=(1-p)\|\mathbf{x}\|_2^{2}$. SPR instead uses a \emph{deterministic} mask that excludes the smallest squared entries of the stacked target from the penalty as long as their cumulative energy is at most $\delta$ (Section~\ref{sec:spr}); writing $\mathcal{M}_d$ for its entries on $\mathbf{x}$ gives $\Omega_{\text{SPR}}(\mathbf{x})=\sum_{d:\,\mathcal{M}_d=0}x_d^{2}$. Thus SPR masks only the penalty, whereas Dropout randomly masks the forward representation.

\textbf{Capacity comparison.}
Both methods use the full embedding at test time, so the capacity is governed by $\|\mathbf{x}\|_2$. Let $\mathcal{F}_B$ be the class of scorers whose embeddings satisfy $\|\mathbf{x}\|_2\le B$, and assume $\hat{\mathfrak{R}}(\mathcal{F}_B)\le C B$ for a constant $C$ independent of $B$, as holds for linear scorers with bounded inputs. Since the excluded coordinates carry energy at most $\delta$,
\[
  \|\mathbf{x}\|_2^{2}
  =\Omega_{\text{SPR}}(\mathbf{x})+\sum_{d:\,\mathcal{M}_d=1}x_d^{2}
  \le\Omega_{\text{SPR}}(\mathbf{x})+\delta .
\]
At a common penalty budget $\tau>0$, SPR therefore admits $\|\mathbf{x}\|_2\le\sqrt{\tau+\delta}$, i.e., $\hat{\mathfrak{R}}_{\text{SPR}}\le C\sqrt{\tau+\delta}$, whereas Dropout admits $\|\mathbf{x}\|_2\le\sqrt{\tau/(1-p)}$, i.e., $\hat{\mathfrak{R}}_{\text{drop}}\le C\sqrt{\tau/(1-p)}$. The SPR bound is strictly smaller if and only if
\[
  \delta<\frac{p}{1-p}\,\tau ,
\]
e.g., when $\tau>4\delta$ for $p=0.2$ or $\tau>\delta$ for $p=0.5$. These bounds concern the constraint sets induced by the two penalties rather than trained models, but they show that SPR controls the full norm almost as tightly as a norm penalty while exempting the low-energy coordinates.

\textbf{Optimality of smallest-first masking.}
Among all sets of coordinates whose total energy is at most $\delta$, the $S$ smallest coordinates selected by SPR form a set of maximum size and, among sets of that size, of minimum energy. Indeed, any $k$ coordinates have at least the energy of the $k$ smallest ones, which exceeds $\delta$ for every $k>S$ by the definition of $S$. SPR thus exempts as many low-energy coordinates as possible while changing the penalty by at most $\delta$ (Lemma~1). Dropout, in contrast, zeroes each coordinate with probability $p$ regardless of its magnitude, so high-energy coordinates are also removed from the forward pass.

\textbf{Example.}
For a target with $k$ coordinates of magnitude $a$ and $n-k$ coordinates of magnitude $\epsilon<a$, any $\delta$ with $(n-k)\epsilon^{2}\le\delta<(n-k)\epsilon^{2}+a^{2}$ makes SPR exempt all low-energy coordinates and penalize only the dominant ones, with penalty $ka^{2}$, while the forward pass still uses the full vector. Dropout removes on average $pk$ dominant coordinates from the score, and its expected penalty $(1-p)\bigl(ka^{2}+(n-k)\epsilon^{2}\bigr)$ is spread over all coordinates.

\textbf{Empirical corroboration.}
On FB15K-237, Dropout raises the MRR of CompGCN-T and CompGCN-C from 0.216 and 0.302 to 0.333 and 0.352, and SPR raises it further to 0.344 and 0.359 (Figure~\ref{fig:delta-gnn}(d)), which is consistent with the analysis above.

\section{SPR Pseudocode}
\label{app:spr-pseudocode}

Algorithm~\ref{alg:spr} gives the implementation-level pseudocode of SPR for a mini-batch. Each sort costs $O(n\log n)$ time, and the mask is not differentiated.

\begin{breakablealgorithm}
\caption{SPR Regularization for a Mini-Batch}
\label{alg:spr}
\begin{algorithmic}[1]
\STATE \textbf{Input:} mini-batch $B$ with stacked embeddings $\mathbf{h}_B,\mathbf{r}_B,\mathbf{t}_B\in\mathbb{R}^{n}$, $n=|B|D$; threshold $\delta > 0$
\STATE \textbf{Output:} batch regularization term $\Omega_{\text{SPR}}$
\STATE $\text{total} \gets 0$
\STATE $\mathbf{x}_1 \gets \mathbf{h}_B^2$; \ $\mathbf{x}_2 \gets \mathbf{t}_B^2$ \COMMENT{Entity terms}
\STATE $\mathbf{x}_3 \gets \mathbf{h}_B^2 \odot \mathbf{r}_B^2$; \ $\mathbf{x}_4 \gets \mathbf{t}_B^2 \odot \mathbf{r}_B^2$ \COMMENT{Interaction terms}
\FOR{$m = 1$ to $4$}
    \STATE Sort $\mathbf{x}_m$ in non-decreasing order: $x_{(1)} \leq \cdots \leq x_{(n)}$
    \STATE $S \gets$ largest index with $\sum_{i=1}^{S} x_{(i)} \leq \delta$ \COMMENT{Masked entries}
    \STATE $\text{total} \gets \text{total} + \sum_{i=S+1}^{n} x_{(i)}$ \COMMENT{Penalize retained entries}
\ENDFOR
\RETURN $\Omega_{\text{SPR}} = \text{total} / |B|$
\end{algorithmic}
\end{breakablealgorithm}

\section{Model Complexity and Generalization}
\textbf{Proposition 1.} \textit{Let \(\mathbf{x}\in\mathbb{R}^n\), and let \(\mathcal{M}=\mathcal{M}(\mathbf{x}^{2})\) be the SPR mask with threshold \(\delta\). Then (i) every retained coordinate is at least as large in magnitude as every masked one, i.e., \(|x_d|\ge|x_{d'}|\) whenever \(\mathcal{M}_d=0\) and \(\mathcal{M}_{d'}=1\); (ii) with the mask held fixed, masked coordinates receive no penalty gradient, \(\partial\Omega_{\text{SPR}}/\partial x_{d'}=0\), while retained coordinates receive \(2x_d\); and (iii) \(\|\mathbf{x}\|_2^2-\delta\le\Omega_{\text{SPR}}(\mathbf{x})\le\|\mathbf{x}\|_2^2\).}

\textbf{Proof.} (i) The mask flags the first \(S\) indices in the non-decreasing order of the \(x_d^2\), so every retained coordinate comes later in this order. (ii) With \(\mathcal{M}\) fixed, \(\Omega_{\text{SPR}}=\sum_{d}x_d^2(1-\mathcal{M}_d)\). (iii) follows from Lemma~1, since \(\|\mathbf{x}\|_2^2-\Omega_{\text{SPR}}(\mathbf{x})=\sum_{d:\mathcal{M}_d=1}x_d^2\).\hfill$\square$

By (iii), SPR bounds the embedding norm, and hence the capacity bound of Appendix~\ref{app:capacity}, almost as a norm penalty does; by (i) and (ii), this control acts only on the largest coordinates. Unlike L2 \cite{hastie2009elements}, SPR leaves the low-energy coordinates, which may carry weak but useful signal, free of the penalty, and unlike Dropout it never perturbs the representation. This targeted capacity control is what distinguishes SPR from L2 and Dropout.

\section{Optimization Dynamics}
The mask \(\mathcal{M}(\mathbf{x})\) is piecewise constant: it changes only where two squared entries tie across the mask boundary or a partial sum equals \(\delta\), which is a set of measure zero. Elsewhere \(\Omega_{\text{SPR}}\) is differentiable with the gradient \(2x_d(1-\mathcal{M}_d(\mathbf{x}))\) of Proposition~1(ii), which is exactly the gradient used in our implementation, since we do not differentiate through the sorting. L2 regularization instead applies the gradient \(2x_d\) to every coordinate, so a gradient step shrinks all coordinates by the same factor \cite{bishop2006pattern}. SPR thus spends the penalty only on the dominant coordinates, which most inflate the embedding norm, while the masked coordinates are updated by the main task loss alone. Moreover, since the SPR mask is a deterministic function of the current embedding, it injects no sampling noise into the gradient, unlike Dropout's random masking \cite{srivastava2014dropout}, which can make stochastic gradient updates more stable \cite{bottou2018optimization}.

\section{Datasets and Experimental Details}
\label{app:datasets}\label{app:setup}
\begin{table}[!tb]
\centering
\caption{Dataset statistics. Ent. and Rel. denote numbers of entities and relations. ICEWS14 spans 2014 (365 timestamps) and ICEWS05-15 spans 2005--2015 (4,017 timestamps); both use 24-hour granularity and the interpolation setting.}
\label{tab:dataset-1}\label{tab:dataset-2}
\footnotesize
\setlength{\tabcolsep}{2pt}
\begin{adjustbox}{max width=\columnwidth}
\begin{tabular}{@{}lrrrrr@{}}
\toprule
Dataset & Ent. & Rel. & Train & Valid & Test \\
\midrule
WN18RR & 40,943 & 11 & 86,835 & 3,034 & 3,314 \\
FB15K-237 & 14,541 & 237 & 272,115 & 17,535 & 20,466 \\
YAGO3-10 & 123,182 & 37 & 1,079,040 & 5,000 & 5,000 \\
Kinship & 104 & 26 & 8,544 & 1,068 & 1,074 \\
UMLS & 135 & 46 & 5,216 & 652 & 661 \\
\midrule
ICEWS14 & 6,869 & 230 & 72,826 & 8,941 & 8,963 \\
ICEWS05-15 & 10,094 & 251 & 368,962 & 46,275 & 46,092 \\
\midrule
WN9-IMG & 6,555 & 9 & 11,741 & 1,337 & 1,319 \\
\bottomrule
\end{tabular}
\end{adjustbox}
\end{table}
Table~\ref{tab:dataset-1} reports the statistics of all datasets. WN18RR \cite{dettmers2018convolutional} and FB15K-237 \cite{toutanova2015observed} are subsets of WordNet and Freebase from which the inverse relations that leak test triples into training have been removed. YAGO3-10 \cite{mahdisoltani2014yago3} is a large-scale subset of YAGO3 in which every entity has at least ten relations. Kinship \cite{kemp2006learning} describes kinship relations among the members of an Australian tribe, and UMLS \cite{mccray2003upper} contains biomedical concepts and their relations from the Unified Medical Language System; both are small. For temporal KGE, ICEWS14 and ICEWS05-15 \cite{garcia2018learning} contain timestamped political events from the Integrated Crisis Early Warning System. For multimodal KGE (Section~\ref{sec:mmkge}), WN9-IMG \cite{xie2016image} is a subset of WN18 whose entities are associated with images.

\paragraph{Implementation details} All experiments were conducted on a server equipped with 1 TB of RAM, an Intel(R) Xeon(R) Gold 6226R CPU @ 2.90GHz, and 8 V100 GPUs with 32 GB each.
All baseline models were reproduced from the GitHub of their respective original papers (including DURA\footnote{\url{https://github.com/MIRALab-USTC/KGE-DURA}}, ER\footnote{\url{https://github.com/Lion-ZS/ER}}, GIE\footnote{\url{https://github.com/Lion-ZS/GIE}}, HGE\footnote{\url{https://github.com/NacyNiko/HGE}}, CompGCN\footnote{\url{https://github.com/malllabiisc/CompGCN}}, and other models) and run on the same machine with the backbone hyperparameters reported in those papers, which all regularizers of the same backbone share (Section~\ref{sec:setting}). For temporal KGE, we use HGE and change only its entity--relation regularization term, keeping its temporal regularizer. All reported results are averaged over three runs, and each run is evaluated at the epoch with the best validation performance.

\paragraph{Hyperparameters} The hyperparameters were obtained from a partial parameter search. Specifically, the parameter $\delta$ was selected from the set $\{0.05, 0.1, 0.2, 0.3, 0.4, 0.45, 0.5\}$, the regularization rate was tuned from the candidate set $\{5 \times 10^{-1}, 5 \times 10^{-2}, 5 \times 10^{-3}, 5 \times 10^{-4}, 5 \times 10^{-5}\}$ (except on WN18RR, where it was tuned from the corresponding $\{1 \times 10^{-1}, 1 \times 10^{-2}, 1 \times 10^{-3}, 1 \times 10^{-4}, 1 \times 10^{-5}\}$ scale), and the embedding dimension was chosen from the range $\{50, 100, 400, 1000, 2000, 3000, 4000\}$. The final hyperparameters were selected according to validation performance.

\paragraph{Evaluation metrics} For each test triple $(h,r,t)$, we answer both the tail query $(h,r,?)$ and the head query $(?,r,t)$. Following the standard filtered setting \cite{bordes2013translating}, the correct entity is ranked against all entities after removing the other candidates that form true triples in the training, validation, or test set. MRR, MR, and Hits@$N$ summarize these ranks:
\begin{equation}\label{eq:metrics}
\begin{gathered}
\mathrm{MRR}=\frac{1}{|S|}\sum_{i\in S}\frac{1}{r_i},\qquad
\mathrm{MR}=\frac{1}{|S|}\sum_{i\in S}r_i,\\
\mathrm{Hits}@N=\frac{1}{|S|}\sum_{i\in S}\mathbb{I}\left(r_i\le N\right),
\end{gathered}
\end{equation}
where $S$ is the set of test queries, $r_i$ is the filtered rank of the correct entity for the $i$-th query, and $\mathbb{I}(\cdot)$ is the indicator function. Higher MRR and Hits@$N$, and lower MR, indicate better performance.

\section{Extended Related Work}
\label{app:related}

\paragraph{Translation-based Models} These models \cite{cao2024knowledge} treat each relation as a translation or geometric transformation between entity vectors. TransE \cite{bordes2013translating} models a fact $\left(e_i, r_k, e_j\right)$ as $e_i + r_k\approx e_j$ and scores it by a distance such as the Euclidean distance. TransH \cite{wang2014knowledge} projects entities onto relation-specific hyperplanes before the translation, and TransR \cite{lin2015learning} maps them into relation-specific spaces. RotatE \cite{sun2019rotate} models relations as rotations in complex space, and HAKE \cite{zhang2020learning} maps entities into polar coordinates to model semantic hierarchies. Other models exploit non-Euclidean geometry: ROTH \cite{chami2020low} applies relation-specific rotations in hyperbolic space, ConE \cite{bai2021modeling} models heterogeneous hierarchies with relation-specific hyperbolic cones, and GIE \cite{cao2022geometry} lets Euclidean, hyperbolic, and hyperspherical spaces interact.

\paragraph{Tensor-decomposition-based Models} These methods factorize the binary KG tensor $\mathcal{A}$ of size $|E| \times |R| \times |E|$, whose entry $\mathcal{A}_{ijk}$ is $1$ if the triple $\left(e_i, r_k, e_j\right)$ exists and $0$ otherwise. The CP decomposition \cite{hitchcock1927expression} approximates it by $d$ rank-one tensors, $\mathcal{A} \approx \sum_{f=1}^d \mathbf{h}_f \circ \mathbf{r}_f \circ \mathbf{t}_f$, with head, relation, and tail factors and the outer product $\circ$. RESCAL \cite{nickel2011three} represents each relation by a full matrix in a bilinear score. DistMult \cite{yang2014embedding} restricts this matrix to be diagonal, which can only model symmetric relations, and ComplEx \cite{trouillon2016complex} moves to complex space to handle asymmetric relations. SimplE \cite{kazemi2018simple} couples the two CP embeddings of each entity through inverse relations, and TuckER \cite{balavzevic2019tucker} uses the Tucker decomposition with a core tensor shared by all relations.

\paragraph{GNN-based, Temporal, and Multimodal Models} GNN-based models such as CompGCN \cite{vashishth2019composition} aggregate neighborhood information through message passing and jointly embed entities and relations with composition operators. Temporal KGE models \cite{wang2023survey,cai2022temporal} associate facts with timestamps; HGE \cite{pan2024hge} embeds temporal KGs in a product space of heterogeneous geometric subspaces. Multimodal KGE models \cite{liang2024survey} enrich structural triples with other modalities: IKRL \cite{xie2016image} learns entity representations from images, and OTKGE \cite{NEURIPS2022_ffdb280e} fuses multimodal embeddings via optimal transport.

\paragraph{Regularization Methods} In KGE, the L2 or Frobenius norm used by early tensor-based models \cite{nickel2011three,yang2014embedding} remains a common baseline. N3 \cite{lacroix2018canonical} applies a weighted tensor nuclear $p$-norm ($p=3$) to the factors of explicit triple interactions and makes CP competitive with more complex models. DURA \cite{zhang2020duality} exploits the duality between tensor-factorization models and distance-based models to regularize entity and relation representations, and its extension \cite{wang2022duality} covers semantic matching KGE models more broadly. ER \cite{cao2022er} is an equivariance regularizer for entity and relation representations, and VIR \cite{xiao2024knowledge} regularizes the norms of the intermediate variables involved in computing the predicted tensor. These norm-based regularizers have mainly been evaluated on TDB models.

Other methods act on the data or the training objective. Mixup \cite{xie2023enhance} interpolates training samples to enhance KGE, and KG-Mixup \cite{shomer2023degree} generates synthetic samples for low-degree entities; Shomer et al. analyze its regularizing effect on low-degree entity--relation pairs. Counterfactual augmentation \cite{chang2023counterfactual} enriches the training graph with counterfactual links. KRACL \cite{tan2023kracl} combines graph context modeling with contrastive learning for sparse KGs, HaSa \cite{zhang2024hasa} addresses false negatives with hardness- and structure-aware contrastive training, and NS-KGE \cite{li2021nonsampling} avoids negative sampling by efficiently learning from all negative instances. These approaches are complementary to SPR\textquotesingle s selective parameter penalty.

\paragraph{Empirical Analysis and Evaluation of KGE} Zhang et al.~\cite{zhang2022rethinkinggcn} isolate the effects of graph structure and representation transformations in GCN-based knowledge graph completion. Tiwari et al.~\cite{tiwari2021revisiting} revisit ranking-based evaluation protocols for link prediction. ReliK~\cite{egger2024relik} estimates the local downstream reliability of embeddings from their scores without retraining, and Asuri Krishnan and Rivero~\cite{krishnan2024patterns} empirically assess the inference patterns captured by trained models. Our study instead measures how regularization affects train--validation gaps under a fixed training protocol.

\end{document}